\documentclass[conference]{IEEEtran}

\usepackage{cite}
\usepackage{amsmath,amssymb,amsfonts}
\usepackage{algorithmic}
\usepackage{graphicx}
\usepackage{textcomp}
\usepackage{xcolor}
\usepackage{array}
\usepackage[hidelinks]{hyperref}
\begin{document}

\title{Evaluating Deep Learning Models for African Wildlife Image Classification: From DenseNet to Vision Transformers}

\author{
\IEEEauthorblockN{Lukman Jibril Aliyu}
\IEEEauthorblockA{\textit{Arewa Data Science Academy} \\
Kano, Nigeria \\
lukman.j.aliyu@gmail.com}
\and
\IEEEauthorblockN{Umar Sani Muhammad}
\IEEEauthorblockA{\textit{Azman University} \\
Kano, Nigeria}
\and
\IEEEauthorblockN{Bilqisu Ismail}
\IEEEauthorblockA{\textit{Arewa Data Science Academy} \\
Kano, Nigeria}
\and
\IEEEauthorblockN{Nasiru Muhammad}
\IEEEauthorblockA{\textit{Arewa Data Science Academy} \\
Kano, Nigeria}
\and
\IEEEauthorblockN{Almustapha A. Wakili}
\IEEEauthorblockA{\textit{Towson University} \\
Maryland, USA}
\and
\IEEEauthorblockN{Seid Muhie Yimam}
\IEEEauthorblockA{\textit{Universität Hamburg} \\
Hamburg, Germany}
\and
\IEEEauthorblockN{Shamsuddeen Hassan Muhammad}
\IEEEauthorblockA{\textit{Imperial College London} \\
London, United Kingdom}
\and
\IEEEauthorblockN{Mustapha Abdullahi}
\IEEEauthorblockA{\textit{Arewa Data Science Academy} \\
Kano, Nigeria}
}

\maketitle

\begin{abstract}
Wildlife populations in Africa face severe threats, with vertebrate numbers declining by over 65\% in the past five decades. In response, image classification using deep learning has emerged as a promising tool for biodiversity monitoring and conservation. This paper presents a comparative study of deep learning models for automatically classifying African wildlife images, focusing on transfer learning with frozen feature extractors. Using a public dataset of four species: buffalo, elephant, rhinoceros, and zebra; we evaluate the performance of DenseNet-201, ResNet-152, EfficientNet-B4, and Vision Transformer ViT-H/14. DenseNet-201 achieved the best performance among convolutional networks (67\% accuracy), while ViT-H/14 achieved the highest overall accuracy (99\%), but with significantly higher computational cost, raising deployment concerns. Our experiments highlight the trade-offs between accuracy, resource requirements, and deployability. The best-performing CNN (DenseNet-201) was integrated into a Hugging Face Gradio Space for real-time field use, demonstrating the feasibility of deploying lightweight models in conservation settings. This work contributes to African-grounded AI research by offering practical insights into model selection, dataset preparation, and responsible deployment of deep learning tools for wildlife conservation.
\end{abstract}


\section{Introduction}
Africa’s rich wildlife heritage faces severe challenges from habitat loss, poaching, and climate change. It is estimated that an elephant is poached every 15 minutes in South Africa \cite{henthorne2020elephant}. Monitoring animal populations traditionally relies on labor-intensive field surveys and camera trap image reviews. Advances in artificial intelligence offer promising tools to automate and ease these efforts. 

Image classification using deep learning has emerged as a powerful approach to biodiversity monitoring \cite{Sharma2024}. By automatically identifying species in photographs taken on camera traps, drones, or smartphones, deep learning systems can greatly accelerate data collection and analysis for conservation biologists. In particular, \cite{Norouzzadeh2018} demonstrated that CNN could accurately identify wildlife species in millions of camera trap images, sometimes exceeding human accuracy. Similarly, projects such as Snapshot Serengeti and the iWildCam challenge have illustrated the potential of AI to handle large-scale wildlife image datasets while highlighting the issue of domain shift between different environments \cite{T_n_2024}. Despite these successes, the implementation of such models in African conservation contexts presents unique challenges. Data on African fauna may be limited or imbalanced towards well-photographed species, and models trained in one context (e.g., certain parks or image conditions) often struggle to generalize to new environments \cite{Xu2024}.

This paper explores the application of deep CNN architectures for classifying African wildlife images, with an emphasis on conditions and species relevant to African conservation. Our work is grounded in the \emph{African Wildlife} \cite{africanWildlifeDataset} dataset, which consists of images of four key species (buffalo, elephant, rhinoceros, and zebra) that are important indicators of the health of the savannah ecosystem. We build on our preliminary findings that a DenseNet-based classifier can achieve promising accuracy on this dataset. DenseNet \cite{Huang2017} is known for its densely connected layers that promote feature reuse and mitigate the vanishing gradient problem. We hypothesize that it will be advantageous for learning from relatively small wildlife datasets. Using ImageNet transfer learning, we fine-tune DenseNet to recognize the target species. In this expanded study, we introduce a comparative analysis with other modern architectures (namely ResNet \cite{He2016}, EfficientNet \cite{Tan2019} and Vision Transformer \cite{Dosovitskiy2021}) to validate our choice of model. Furthermore, we significantly extend our review and discussion of the literature to contextualize our contributions within existing efforts in \textit{AI for Social Good} and African sustainability goals.

Our contributions are threefold: 
\begin{itemize}
    \item \textbf{Improved Image Classification Pipeline:} We present an improved image classification pipeline for African wildlife conservation, incorporating tailored dataset preprocessing and model fine-tuning strategies specifically adapted to African ecological contexts.

    \item \textbf{Comparative Evaluation of Architectures:} We conduct a comprehensive evaluation of DenseNet’s performance relative to other deep learning architectures, providing insights into the trade-offs between accuracy and computational efficiency.

    \item \textbf{Ethical and Responsible AI Deployment:} We explicitly address the ethical considerations, limitations, and broader societal impacts of deploying AI systems for conservation in Africa.
\end{itemize}

\section{Related Work}

\paragraph{Deep Learning for Wildlife Conservation.}
The intersection of computer vision and wildlife conservation has attracted considerable interest in recent years. Early pioneering work by \cite{Norouzzadeh2018} applied deep CNNs (e.g., ResNet-152) to the Snapshot Serengeti camera trap dataset \cite{dryad_5pt92}, achieving remarkable accuracy in identifying over 40 species and demonstrating that deep learning can greatly improve ecological data processing. Following this, \cite{Beery2018} emphasized the challenge of generalizing models to new locations (``unseen domains'') in the wildlife context. Their iWildCam 2018 study on the iwildCam dataset \cite{beery2019iwildcam2018challengedataset} showed that models trained on one set of camera traps suffered performance drops when applied to data from different regions, underlining the importance of diverse training data and domain adaptation techniques. More recently, the WILDS benchmark \cite{Koh2021} formalized such distribution shift challenges, including an animal camera trap classification task that tests robustness of models to changes in geographical location and imaging conditions.

Beyond camera trap imagery, deep learning has been used in various conservation scenarios. For instance, \cite{Bothmann2023} developed an active learning framework to improve species classification, reducing the annotation burden by iteratively selecting the most informative wildlife images for labeling. Their approach, evaluated on a large collection of European and African wildlife images, highlights how human-in-the-loop strategies can address dataset bias and scarcity. In aerial and drone-based wildlife monitoring, CNNs and object detectors have been used to detect animals in overhead imagery \cite{Xu2024}, expanding the toolkit for conservationists to include surveillance from the skies. These efforts collectively illustrate the growing role of AI in biodiversity assessment.

\paragraph{CNN Architectures in Image Classification.}
Convolutional neural networks have seen rapid evolution, with numerous architectures pushing the state of the art on image classification benchmarks. ResNet \cite{He2016}, introduced in 2015, demonstrated that very deep networks (with 50+ layers) could be effectively trained using residual skip connections, and it remains a popular backbone for many vision tasks. DenseNet \cite{Huang2017} further innovated by connecting each layer to all subsequent layers, maximizing feature reuse and alleviating vanishing gradients; this compact architecture often achieves comparable accuracy to deeper ResNets with fewer parameters. Another notable family is EfficientNet \cite{Tan2019}, which introduced a principled compound scaling method to balance network depth, width, and resolution, leading to a series of models ( from B0 to B7) that achieved excellent accuracy with high parameter efficiency. In parallel, the vision transformer (ViT) architecture \cite{Dosovitskiy2021} showed that transformer models (prevalent in NLP) can also excel at image recognition when trained on large datasets, by dividing images into patch embeddings and relying on self-attention mechanisms instead of convolutions.

In wildlife image classification tasks, most studies have employed CNN architectures (often ResNet-based) via transfer learning. \cite{Ukwuoma2022} proposed a modified attention-based CNN (incorporating feature pyramid networks) and tested it on the same African Wildlife dataset used in this work, as well as an Animal-80 dataset; their method improved detection and classification performance by leveraging multi-scale features. Another recent study by \cite{Sharma2024} compared several deep models (DenseNet, ResNet, VGG, and the YOLOv8 detector) on a custom wildlife dataset of 23 endangered species. They reported that the YOLOv8 model, though originally designed for object detection, achieved the highest classification accuracy (over 96\% F1-score), outperforming the CNN classifiers. However, simpler CNNs like ResNet or DenseNet still remain competitive baselines, especially in scenarios with limited computational resources or where interpretability of the classification is needed (since one can visualize CNN feature maps more straightforwardly than transformer attention, for example). A survey by \cite{Sultana2019} provides a comprehensive overview of advancements in CNN-based image classification up to 2019, noting trends such as the move towards deeper but more efficient networks and the adoption of transfer learning as a standard practice for limited data scenarios. While many CNN architectures are applicable to wildlife image classification, the choice often depends on the specific context: the size of the dataset, the need for speed (e.g., edge deployment), and the importance of generalization across domains. Our work contributes to this body of knowledge by evaluating DenseNet in an African wildlife context, using a dataset gotten from Africa, and comparing it with a ResNet and EfficientNet baseline, and a vision transformer. We also discuss the ethical deployment of such models in conservation settings, an aspect less frequently addressed in technical studies.

\section{Methodology}

\begin{figure}[htbp]
    \centering
    \includegraphics[width=0.5\textwidth]{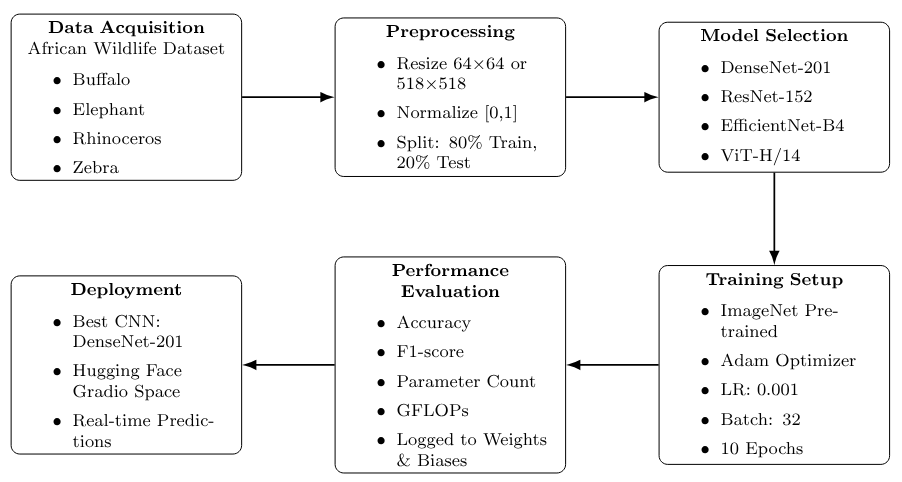}
    \caption{Deep Learning Pipeline for African Wildlife Species Classification. The workflow encompasses all stages from initial data acquisition through final deployment for real-time species identification in conservation applications.}
    \label{fig:methodology_pipeline}
\end{figure}

This study followed a structured pipeline comprising data acquisition, preprocessing, model selection, training, and deployment as illustrated in Figure ~\ref{fig:methodology_pipeline}. Our goal was to evaluate the performance of different deep learning architectures on a balanced wildlife image dataset, with an emphasis on efficient, ethical, and deployable AI for conservation efforts.

\subsection{Dataset and Preprocessing and Models}

In this study, we utilize the publicly available African Wildlife dataset \cite{africanWildlifeDataset}, which consists of color images spanning four animal species: buffalo, elephant, rhinoceros, and zebra. Each class contains 376 images, resulting in a balanced dataset with a total of 1,504 images. To enable robust model evaluation, the dataset is partitioned into two subsets: a training set comprising 1,203 images (80\%) and a test set comprising 301 images (20\%). This split ensures a consistent framework for assessing classification performance across models.

Each image was resized to $64 \times 64$ pixels and normalized to the $[0,1]$ range. These preprocessing steps were adapted from \cite{Shorten2019DataAA}, and they were essential in enhancing model generalization and robustness, especially with the relatively small dataset size. However, for the vision transformer training, the images were resized to $518 \times 518$ pixels as that is the minimum size that the model can take \cite{torchvision-vith14}. 

We evaluated four pretrained deep learning models for the classification task: \textbf{DenseNet-201} \cite{huang2017densely}
,\textbf{EfficientNet-B4} \cite{Tan2019}
,\textbf{ResNet-152} \cite{He2016}
and \textbf{Vision Transformer (ViT-H/14)} \cite{Dosovitskiy2021}. All models were initialized with ImageNet-pretrained weights using \texttt{torchvision.models} \cite{torchvision-models}. To reduce training time and avoid overfitting, we froze all feature extraction layers and fine-tuned only the final classification layers.

\subsection{Training Setup}

All experiments were conducted in the Kaggle cloud environment using an NVIDIA Tesla P100 GPU (16GB VRAM, CUDA 12.6, driver version 560.35.03). The models were implemented in PyTorch, with experiment tracking and visualization performed via Weights \& Biases (W\&B) \cite{wandb}.

Each model was trained using the following configuration:

\begin{itemize}
    \item \textbf{Optimizer:} Adam
    \item \textbf{Learning Rate:} 0.001
    \item \textbf{Loss Function:} CrossEntropyLoss
    \item \textbf{Batch Size:} 32
    \item \textbf{Epochs:} 10
    \item \textbf{Input Resolution:} $64 \times 64$ ($518 \times 518$ for vision transformer as is required by the model)
\end{itemize}

Training durations varied by architecture. DenseNet, EfficientNet and ResNet completed training in under two minutes, while the ViT-H/14 model required over an hour due to its large number of parameters and transformer-based design. Performance metrics, parameter counts, and training times are summarized in Table~\ref{tab:model-stats}.

\subsection{Experiment Tracking}

We used Weights \& Biases (W\&B) \cite{wandb} for experiment tracking, logging, and visualization throughout this study. Key metrics such as training and validation accuracy, loss, F1-score, precision, and recall were monitored in real time using W\&B dashboards. Additionally, GPU power consumption during training was recorded and visualized to compare the computational efficiency of different architectures. All training runs were tagged, versioned, and documented via W\&B to ensure reproducibility and systematic model comparison, and can be viewed at \url{https://wandb.ai/lukmanaj/africa-wildlife-dli?nw=nwuserlukmanaj}

\subsection{Deployment}

To demonstrate real-world applicability, we exported the best-performing convolutional model—DenseNet-201—and deployed it as an interactive web application using Hugging Face Gradio. This application, available at \url{https://huggingface.co/spaces/lukmanaj/afri-wildlife-classify}, enables conservationists and researchers to upload wildlife images and receive species predictions in real time, showcasing the practical potential of AI for biodiversity monitoring and wildlife protection.

\section{Experimental Results}

We evaluated four pretrained models: DenseNet-201, EfficientNet-B4, ResNet-152, and ViT-H/14 on the African Wildlife dataset, which consists of four species classes. Table~\ref{tab:f1-comparison} reports the overall classification accuracy, macro-averaged F1-score, and per-class F1-scores.

Among the CNNs trained, DenseNet-201 achieved 67\% accuracy and the highest F1-scores for the buffalo (0.72) and zebra (0.76) classes. EfficientNet-B4 performed the worst overall, with 48\% accuracy and a macro F1-score of 0.47. ResNet-152 yielded moderate results across all metrics.

ViT-H/14 significantly outperformed all the other models, achieving 99\% accuracy and the highest F1-scores for all classes.

\begin{table}[htbp]
\centering
\caption{Performance of models on the African Wildlife test set. Metrics include accuracy, macro F1-score, and per-class F1-scores.}
\label{tab:f1-comparison}
\setlength{\tabcolsep}{3pt} 
\renewcommand{\arraystretch}{1.1} 
\small 
\begin{tabular}{lcccccc}
\hline
\textbf{Model} & \textbf{Acc.} & \textbf{Macro F1} & \textbf{Buffalo} & \textbf{Elephant} & \textbf{Rhino} & \textbf{Zebra} \\
\hline
DenseNet-201    & 67.0\% & 0.67 & 0.72 & 0.61 & 0.60 & 0.76 \\
EfficientNet-B4 & 48.0\% & 0.47 & 0.54 & 0.47 & 0.40 & 0.48 \\
ResNet-152      & 57.0\% & 0.58 & 0.56 & 0.56 & 0.52 & 0.67 \\
ViT-H/14        & 99.0\% & 0.99 & 0.99 & 0.99 & 0.99 & 0.99 \\
\hline
\end{tabular}
\end{table}

To provide additional context for model selection, Table~\ref{tab:model-stats} summarizes each model's parameter count, estimated GFLOPs, Giga Floating Point Operations Per Second, which is a unit of measurement that describes a computer's processing power, specifically its ability to perform floating-point operations (based on standard input resolution), training time, and notes on deployment feasibility. While ViT-H/14 achieved the highest accuracy, its large compute and memory footprint makes DenseNet-201 a more practical option for lightweight deployments.

\begin{table}[htbp]
\centering
\caption{Model characteristics and training times. GFLOPs based on 224×224 input. Only classification layers were fine-tuned.}
\label{tab:model-stats}
\setlength{\tabcolsep}{1.5pt}
\renewcommand{\arraystretch}{1.1}
\small
\begin{tabular}{l>{\raggedright}p{1.2cm}>{\raggedright}p{1.2cm}>{\raggedright}p{1cm}>{\raggedright}p{1cm}>{\raggedright\arraybackslash}p{2.2cm}}
\hline
\textbf{Model} & \textbf{Params (M)} & \textbf{GFLOPs} & \textbf{Acc.} & \textbf{Time (s)} & \textbf{Notes} \\
\hline
DenseNet-201    & 20.0   & 4.29     & 67\% & 92.5   & Gradio deployment \\
EfficientNet-B4 & 19.3   & 4.39     & 48\% & 87.8   & Under-performed \\
ResNet-152      & 60.2   & 11.51    & 57\% & 83.2   & Good baseline \\
ViT-H/14        & 632.0  & 1016.7   & 99\% & 6574.2 & Resource intensive \\
\hline
\end{tabular}
\end{table}
\section{Discussion}

Our experiments underscore the potential of deep learning for wildlife image classification in African contexts. Using DenseNet-201, we achieved a test accuracy of 67\% and macro F1-score of 0.67, establishing a strong CNN-based baseline across four key species: buffalo, elephant, rhino, and zebra. DenseNet consistently outperformed EfficientNet-B4 and ResNet-152 in our setting, particularly in terms of per-class F1-scores for buffalo and zebra. This supports the hypothesis that DenseNet’s densely connected layers facilitate better feature propagation and reuse, making it well-suited for small and low-resolution datasets often encountered in conservation tasks.

In contrast, EfficientNet-B4 underperformed (48\% accuracy), despite its strong performance on ImageNet benchmarks. This may be due to its compound scaling design, which can be sensitive to input resolution and small datasets. ResNet-152 achieved moderate results (57\% accuracy), validating its robustness but still falling short of DenseNet in this task.

As shown in Appendix~\ref{fig:gpu-power}, the Vision Transformer consumed significantly more GPU power throughout training compared to the CNN models. Detailed performance metrics, including macro-averaged F1 (\ref{fig:macro-f1}), precision (\ref{fig:macro-precision}), and recall (\ref{fig:macro-recall}) scores, are provided in the appendix.

\paragraph{CNNs vs. Vision Transformers.}
The most striking result came from the Vision Transformer ViT-H/14, which achieved 99\% accuracy and near-perfect precision and recall across all classes. This highlights the potential of transformer-based models in wildlife classification tasks—particularly when leveraging large-scale pretraining. However, ViT-H/14 has over 600M parameters and a high computational footprint, making it unsuitable for deployment on low-resource or edge devices without further model compression or distillation \cite{saha2025visiontransformersedgecomprehensive}. This is quite important if it will be integrated into a system that does retraining using human-in-the-loop, as it will be computationally expensive to constantly retrain the model. Furthermore, in cases of use in places without good internet connection, performing inference offline will be faster in a lightweight model.

By comparison, CNNs like DenseNet offer a more favorable balance between accuracy and deployability. While they may not match ViT-level accuracy, they can still provide robust performance at a fraction of the compute cost, especially if they can be improved.

\paragraph{Deployment and Domain Shift.}
Our deployed prototype is developed using a fine-tuned DenseNet-201 model and implemented as a Hugging Face Gradio Space. This app enables field users to upload images and receive species predictions in real-time. However, when tested on smartphone-captured field images, performance declined sharply. This domain shift between curated training data and real-world imagery is well documented by \cite{Beery2018} and \cite{Koh2021}. It emphasizes the need for more diverse, representative training datasets, incorporating variations in lighting, background, camera angle, and image quality.

\paragraph{Comparison to Prior Work.}
Our approach is relatively lightweight compared to detection-based or attention-enhanced models. Prior studies such as \cite{Ukwuoma2022} and \cite{Sharma2024} have shown that YOLOv8 and similar architectures can yield higher accuracy by combining classification with object localization. While ViTs partially address this through attention, dedicated detection frameworks remain an attractive next step for improving real-world performance.

\paragraph{Concluding Insights.}
Despite limitations, our work illustrates that end-to-end AI tools for conservation are feasible using accessible tools and public data. Unlike many studies that stop at offline accuracy, we demonstrate a functioning pipeline from training to deployment, making our research an impactful, Africa-grounded machine learning and providing a foundation for future work on data diversity, model robustness, and real-world usability.

\section{Conclusion and Future Work}
In this paper, we presented a deep learning approach for classifying African wildlife images, using DenseNet-201 model as the primary architecture. Our model achieved a test accuracy of 67\% on a four-species dataset, demonstrating the feasibility of CNNs for species recognition with limited data. We also deployed the model in a functional Hugging Face Gradio Space for user-friendly interaction.

Future work will focus on expanding the dataset—both in size and species diversity—through collaborations and possibly integrating camera trap images like those from Snapshot Serengeti \cite{Norouzzadeh2018}. Advanced data augmentation or synthetic data generation (e.g., with GANs) may further improve model robustness.

We also plan to enhance the deployed app by incorporating user feedback for active learning and exploring deployment on edge devices (e.g., NVIDIA Jetson Nano) for offline use in remote areas \cite{Ingaleshwar2024}. Ethical AI practices by \cite{WWF2022} will guide our work, and we aim to release our dataset and code publicly to support open, Africa-centric AI research.
Finally, we plan to test the system under diverse African field conditions and on edge devices, to ensure the model’s effectiveness in real conservation deployments. 

\section{Ethical Considerations and Limitations}

Bias in the dataset favoring well-photographed conditions could lead to uneven model performance across species or environments. Ethical use requires transparency about such limitations, especially when models may underperform on rarer or nocturnal species. Furthermore, data provenance must be considered; although the Kaggle dataset was public, deployment may require further permissions.

Privacy concerns also arise if human subjects are unintentionally captured in future datasets. Any scale-up involving camera traps should include privacy safeguards. Deployment risks such as over-reliance and potential misuse (e.g., by poachers) necessitate a human-in-the-loop approach and access controls.

Overall, the current model is a proof-of-concept and not yet robust enough for critical decision-making; however, it demonstrates value as a prototype for ranger-assisted monitoring and citizen science.

\section{Broader Impact}

This work contributes to the \textit{AI for Social Good} agenda by applying machine learning to biodiversity monitoring, supporting SDG 15 (\emph{Life on Land}). Automatic classification can accelerate wildlife surveys, improve anti-poaching efforts, and unlock underused camera trap datasets.

Our African-centered approach demonstrates that impactful AI research can emerge from local challenges. By building and sharing open tools, we promote capacity-building among African researchers and ecologists. The methodology can also inspire adjacent applications such as crop monitoring or disease surveillance.

Potential negative impacts, such as overreliance on AI or misuse, are acknowledged but can be mitigated through careful design and stakeholder participation. This project lays the groundwork for responsible, locally grounded AI systems that improve ecological conservation.

\section*{Acknowledgment}

We thank Arewa Data Science Academy for supporting this project through the Arewa Data Science Deep Learning with PyTorch fellowship. We are also grateful to the developers of the African Wildlife Dataset on Kaggle (Bianca Ferreira) and the open-source deep learning community.

\bibliographystyle{IEEEtran}
\bibliography{sample}

\begin{thebibliography}{10}
\providecommand{\url}[1]{#1}
\csname url@samestyle\endcsname
\providecommand{\newblock}{\relax}
\providecommand{\bibinfo}[2]{#2}
\providecommand{\BIBentrySTDinterwordspacing}{\spaceskip=0pt\relax}
\providecommand{\BIBentryALTinterwordstretchfactor}{4}
\providecommand{\BIBentryALTinterwordspacing}{\spaceskip=\fontdimen2\font plus
\BIBentryALTinterwordstretchfactor\fontdimen3\font minus \fontdimen4\font\relax}
\providecommand{\BIBforeignlanguage}[2]{{%
\expandafter\ifx\csname l@#1\endcsname\relax
\typeout{** WARNING: IEEEtran.bst: No hyphenation pattern has been}%
\typeout{** loaded for the language `#1'. Using the pattern for}%
\typeout{** the default language instead.}%
\else
\language=\csname l@#1\endcsname
\fi
#2}}
\providecommand{\BIBdecl}{\relax}
\BIBdecl

\bibitem{henthorne2020elephant}
\BIBentryALTinterwordspacing
P.~Henthorne, ``Elephant poaching in south africa,'' May 2020, university of San Francisco Office of Sustainability – Student Blog. [Online]. Available: \url{https://usfblogs.usfca.edu/sustainability/2020/05/15/elephant-poaching-in-south-africa/}
\BIBentrySTDinterwordspacing

\bibitem{Sharma2024}
S.~Sharma, S.~Dhakal, and M.~Bhavsar, ``Transfer learning for wildlife classification: Evaluating {YOLOv8} against densenet, resnet, and vggnet on a custom dataset,'' \emph{Journal of Artificial Intelligence and Capsule Networks}, vol.~6, no.~4, pp. 415--435, 2024.

\bibitem{Norouzzadeh2018}
M.~S. Norouzzadeh, A.~Nguyen, M.~Kosmala, A.~Swanson, C.~Packer, and J.~Clune, ``Automatically identifying, counting, and describing wild animals in camera-trap images with deep learning,'' \emph{Proceedings of the National Academy of Sciences (PNAS)}, vol. 115, no.~25, pp. E5716--E5725, 2018.

\bibitem{T_n_2024}
\BIBentryALTinterwordspacing
A.~Tøn, A.~Ahmed, A.~S. Imran, M.~Ullah, and R.~M.~A. Azad, ``Metadata augmented deep neural networks for wild animal classification,'' \emph{Ecological Informatics}, vol.~83, p. 102805, Nov. 2024. [Online]. Available: \url{http://dx.doi.org/10.1016/j.ecoinf.2024.102805}
\BIBentrySTDinterwordspacing

\bibitem{Xu2024}
Z.~Xu, T.~Wang, A.~K. Skidmore, S.~D. Phinn, and L.~Wang, ``A review of deep learning techniques for detecting animals in aerial and satellite images,'' \emph{International Journal of Applied Earth Observation and Geoinformation}, vol. 128, p. 103732, 2024.

\bibitem{africanWildlifeDataset}
B.~Ferreira, ``African wildlife dataset,'' \url{https://www.kaggle.com/datasets/biancaferreira/african-wildlife/data}, 2020, accessed: 2024-02-13.

\bibitem{Huang2017}
G.~Huang, Z.~Liu, L.~V.~D. Maaten, and K.~Q. Weinberger, ``Densely connected convolutional networks,'' in \emph{Proc. IEEE Conference on Computer Vision and Pattern Recognition (CVPR)}, 2017, pp. 4700--4708.

\bibitem{He2016}
K.~He, X.~Zhang, S.~Ren, and J.~Sun, ``Deep residual learning for image recognition,'' in \emph{Proc. IEEE Conference on Computer Vision and Pattern Recognition (CVPR)}, 2016, pp. 770--778.

\bibitem{Tan2019}
M.~Tan and Q.~V. Le, ``Efficientnet: Rethinking model scaling for convolutional neural networks,'' in \emph{Proc. International Conference on Machine Learning (ICML)}, ser. PMLR, vol.~97, 2019, pp. 6105--6114.

\bibitem{Dosovitskiy2021}
A.~Dosovitskiy, L.~Beyer, A.~Kolesnikov, D.~Weissenborn, and et~al., ``An image is worth 16x16 words: Transformers for image recognition at scale,'' in \emph{Proc. International Conference on Learning Representations (ICLR)}, 2021.

\bibitem{dryad_5pt92}
\BIBentryALTinterwordspacing
A.~Swanson, M.~Kosmala, C.~Lintott, R.~Simpson, A.~Smith, and C.~Packer, ``Data from: Snapshot serengeti, high-frequency annotated camera trap images of 40 mammalian species in an african savanna,'' 2015. [Online]. Available: \url{https://doi.org/10.5061/dryad.5pt92}
\BIBentrySTDinterwordspacing

\bibitem{Beery2018}
S.~Beery, G.~V. Horn, and P.~Perona, ``Recognition in terra incognita: Wildlife object classification in unseen domains,'' in \emph{Proc. European Conference on Computer Vision (ECCV) Workshops}, 2018, pp. 52--68.

\bibitem{beery2019iwildcam2018challengedataset}
\BIBentryALTinterwordspacing
S.~Beery, G.~van Horn, O.~M. Aodha, and P.~Perona, ``The iwildcam 2018 challenge dataset,'' 2019. [Online]. Available: \url{https://arxiv.org/abs/1904.05986}
\BIBentrySTDinterwordspacing

\bibitem{Koh2021}
P.~W. Koh, S.~Sagawa, H.~Marklund, and et~al., ``{WILDS}: A benchmark of in-the-wild distribution shifts,'' in \emph{Proc. International Conference on Machine Learning (ICML)}, ser. PMLR, vol. 139, 2021, pp. 5637--5664.

\bibitem{Bothmann2023}
L.~Bothmann, L.~Wimmer, O.~Charrakh, T.~Weber, H.~Edelhoff, and W.~Peters, ``Automated wildlife image classification: An active learning tool for ecological applications,'' \emph{Ecological Informatics}, vol.~77, p. 102231, 2023.

\bibitem{Ukwuoma2022}
C.~C. Ukwuoma, Z.~guang Qin, G.~U. Nneji, and G.~C. Urama, ``Animal species detection and classification framework based on modified multi-scale attention mechanism and feature pyramid network,'' \emph{Scientific African}, vol.~16, p. e01151, 2022.

\bibitem{Sultana2019}
F.~Sultana, A.~Sufian, and P.~Dutta, ``Advancements in image classification using convolutional neural network,'' \emph{arXiv preprint arXiv:1905.03288}, 2019.

\bibitem{Shorten2019DataAA}
C.~Shorten and T.~M. Khoshgoftaar, ``A survey on data augmentation for deep learning,'' \emph{Journal of Big Data}, vol.~6, p.~60, 2019.

\bibitem{torchvision-vith14}
\BIBentryALTinterwordspacing
P.~C. Team, ``torchvision.models.vit\_h\_14,'' 2025, accessed: 2025-05-13. [Online]. Available: \url{https://docs.pytorch.org/vision/main/models/generated/torchvision.models.vit_h_14.html}
\BIBentrySTDinterwordspacing

\bibitem{huang2017densely}
G.~Huang, Z.~Liu, L.~Van Der~Maaten, and K.~Q. Weinberger, ``Densely connected convolutional networks,'' \emph{Proceedings of the IEEE conference on computer vision and pattern recognition}, pp. 4700--4708, 2017.

\bibitem{torchvision-models}
{Torchvision Contributors}, ``Models and pre-trained weights,'' \url{https://pytorch.org/vision/main/models.html}, accessed: 2025-05-18.

\bibitem{wandb}
\BIBentryALTinterwordspacing
L.~Biewald, ``Experiment tracking with weights and biases,'' 2020, software available from wandb.ai. [Online]. Available: \url{https://www.wandb.ai/}
\BIBentrySTDinterwordspacing

\bibitem{saha2025visiontransformersedgecomprehensive}
\BIBentryALTinterwordspacing
S.~Saha and L.~Xu, ``Vision transformers on the edge: A comprehensive survey of model compression and acceleration strategies,'' 2025. [Online]. Available: \url{https://arxiv.org/abs/2503.02891}
\BIBentrySTDinterwordspacing

\bibitem{Ingaleshwar2024}
S.~Ingaleshwar, F.~Tasharofi, M.~A. Pava, H.~Vaishya, and et~al., ``Wildlife species classification on the edge: A deep learning perspective,'' in \emph{Proc. 16th Int. Conf. on Agents and Artificial Intelligence (ICAART)}, 2024, pp. 600--608.

\bibitem{WWF2022}
{World Wide Fund for Nature (WWF)}, ``Living planet report 2022 -- regional fact sheet: Africa,'' \url{https://africa.panda.org/factsheets/}, 2022, accessed 2025-05-09.

\end{thebibliography}

\clearpage
\appendix
\section*{Additional Training Statistics}
\label{add:training-stats}
\begin{figure}[htbp]
    \centering
    \includegraphics[width=0.5\textwidth]{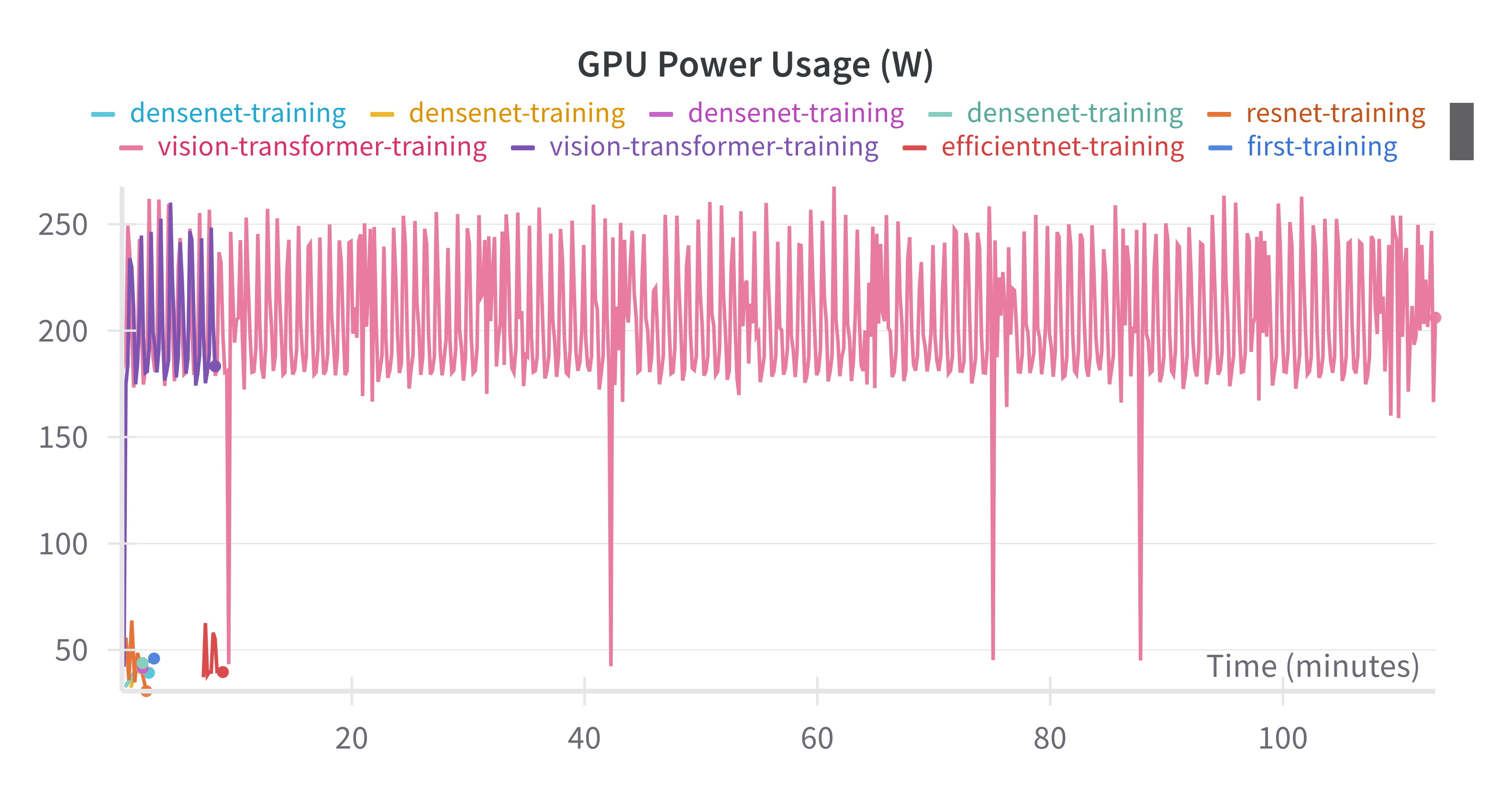}
    \caption{GPU power usage during model training.}
    \label{fig:gpu-power}
\end{figure}

\begin{figure}[htbp]
    \centering
    \includegraphics[width=0.5\textwidth]{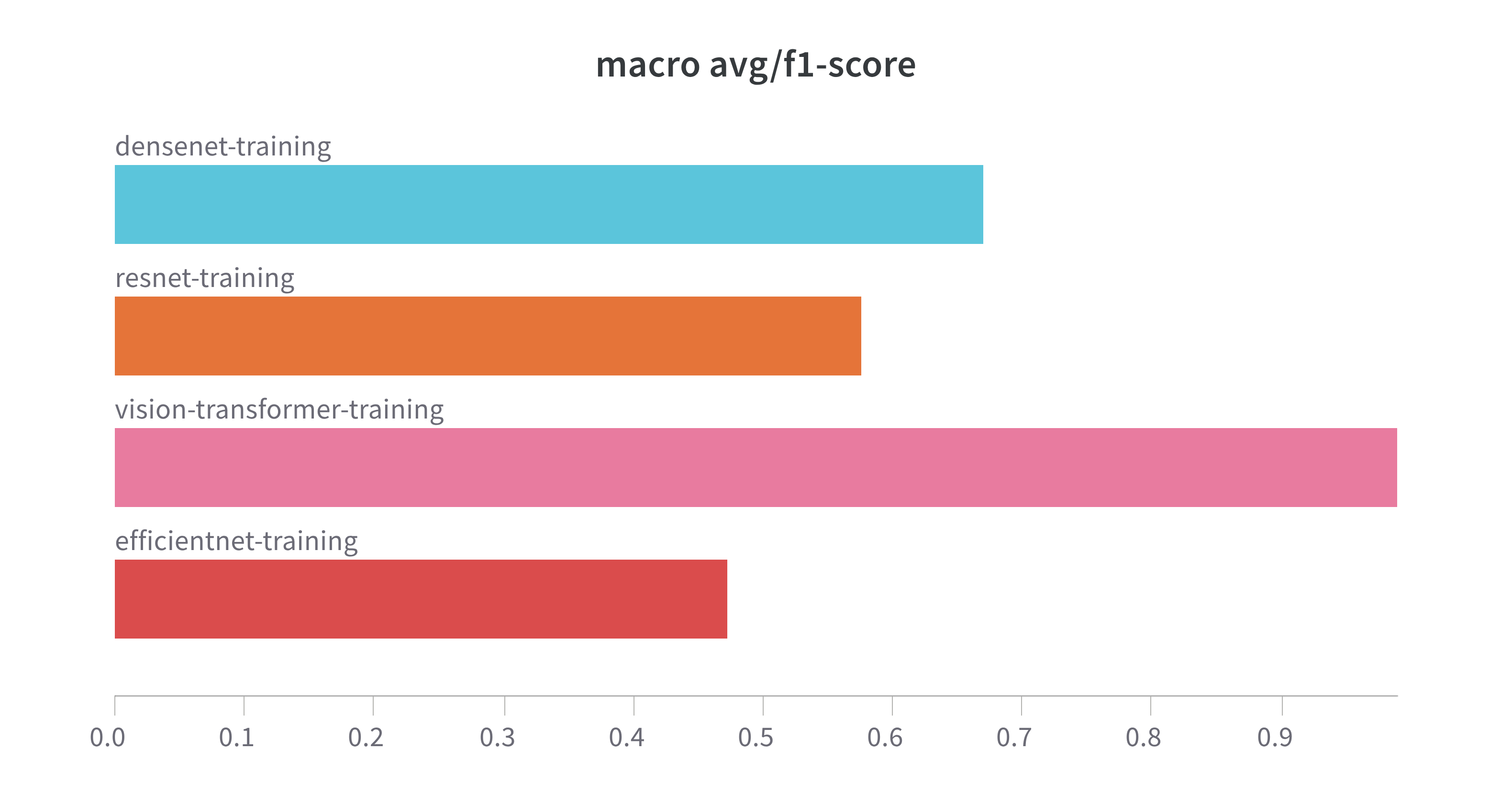}
    \caption{Macro-averaged F1-score for each model.}
    \label{fig:macro-f1}
\end{figure}

\begin{figure}[htbp]
    \centering
    \includegraphics[width=0.5\textwidth]{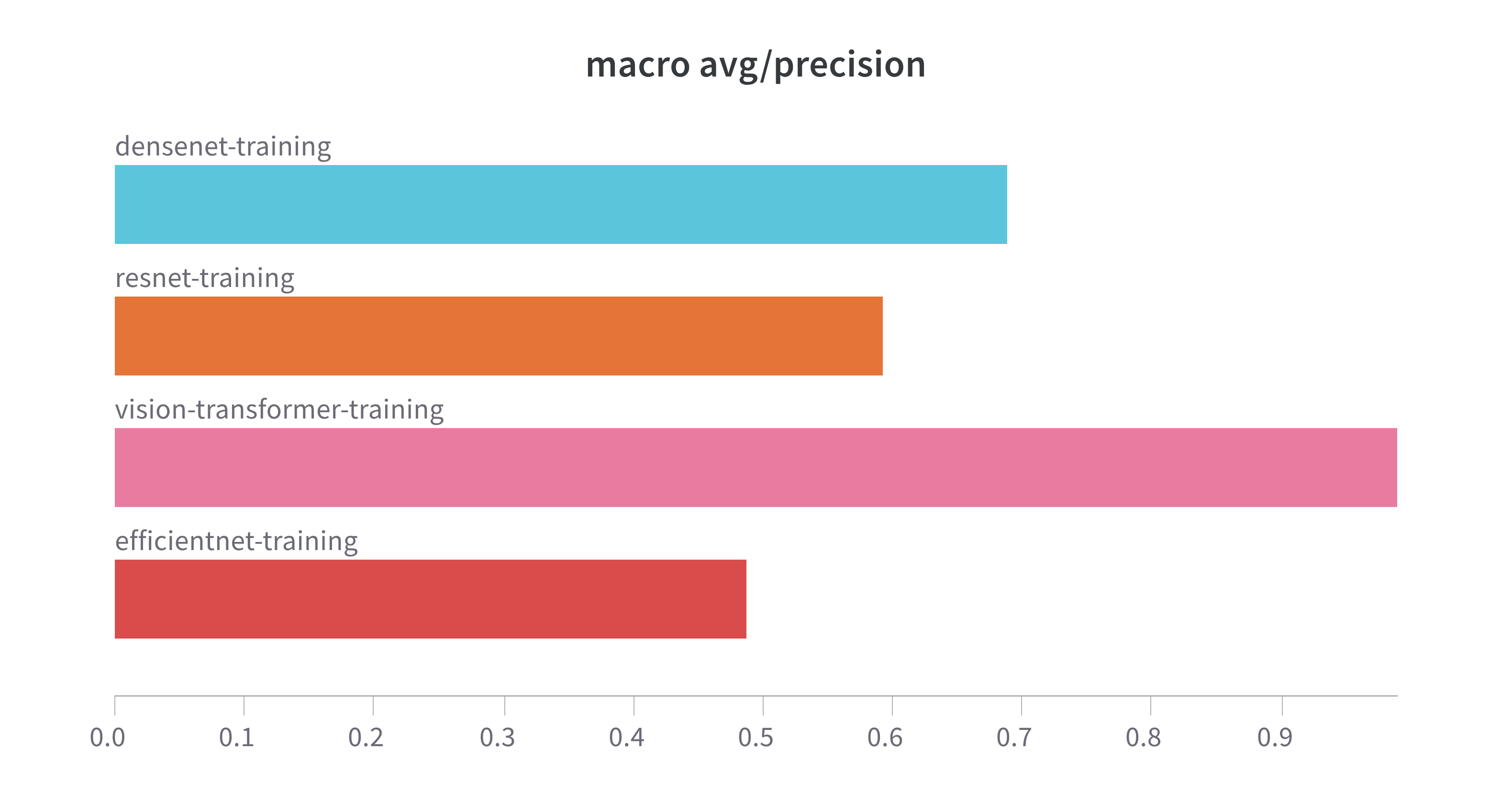}
    \caption{Macro-averaged precision scores.}
    \label{fig:macro-precision}
\end{figure}

\begin{figure}[htbp]
    \centering
    \includegraphics[width=0.5\textwidth]{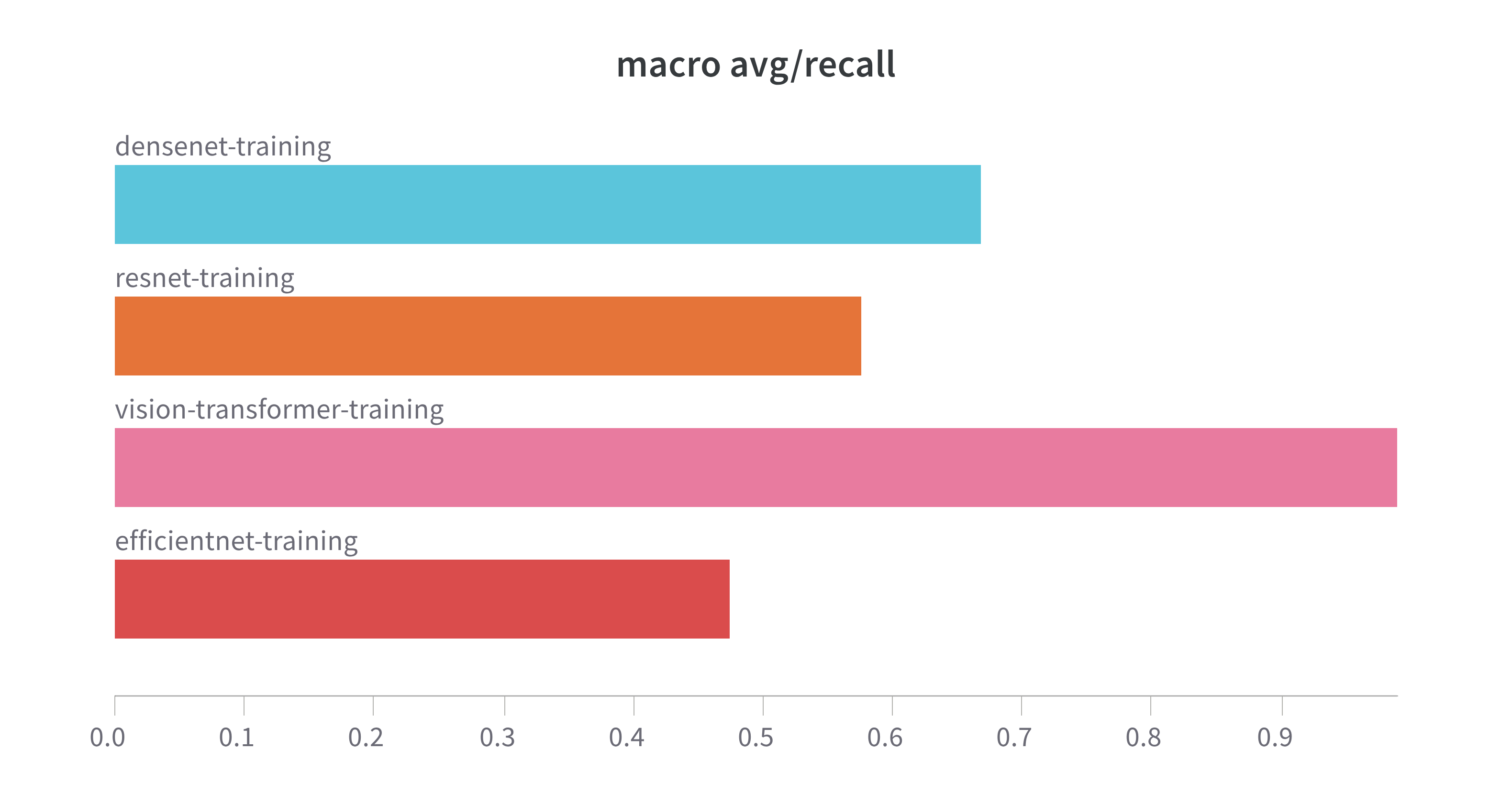}
    \caption{Macro-averaged recall scores.}
    \label{fig:macro-recall}
\end{figure}

\begin{figure}[htbp]
    \centering
    \includegraphics[width=0.5\textwidth]{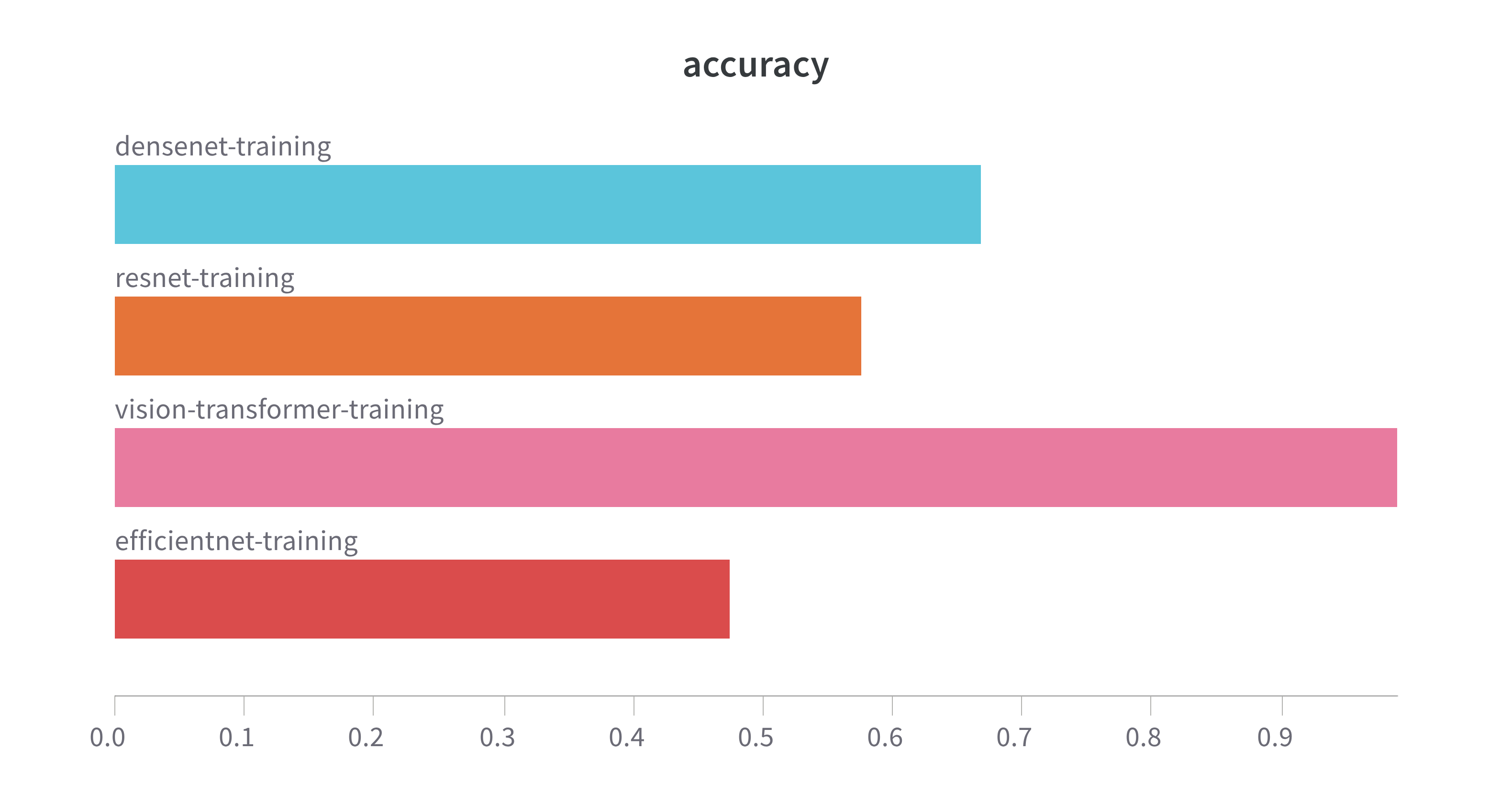}
    \caption{Overall accuracy comparison.}
    \label{fig:accuracy}
\end{figure}

\newpage

\begin{figure}[htbp]
    \centering
    \includegraphics[width=0.5\textwidth]{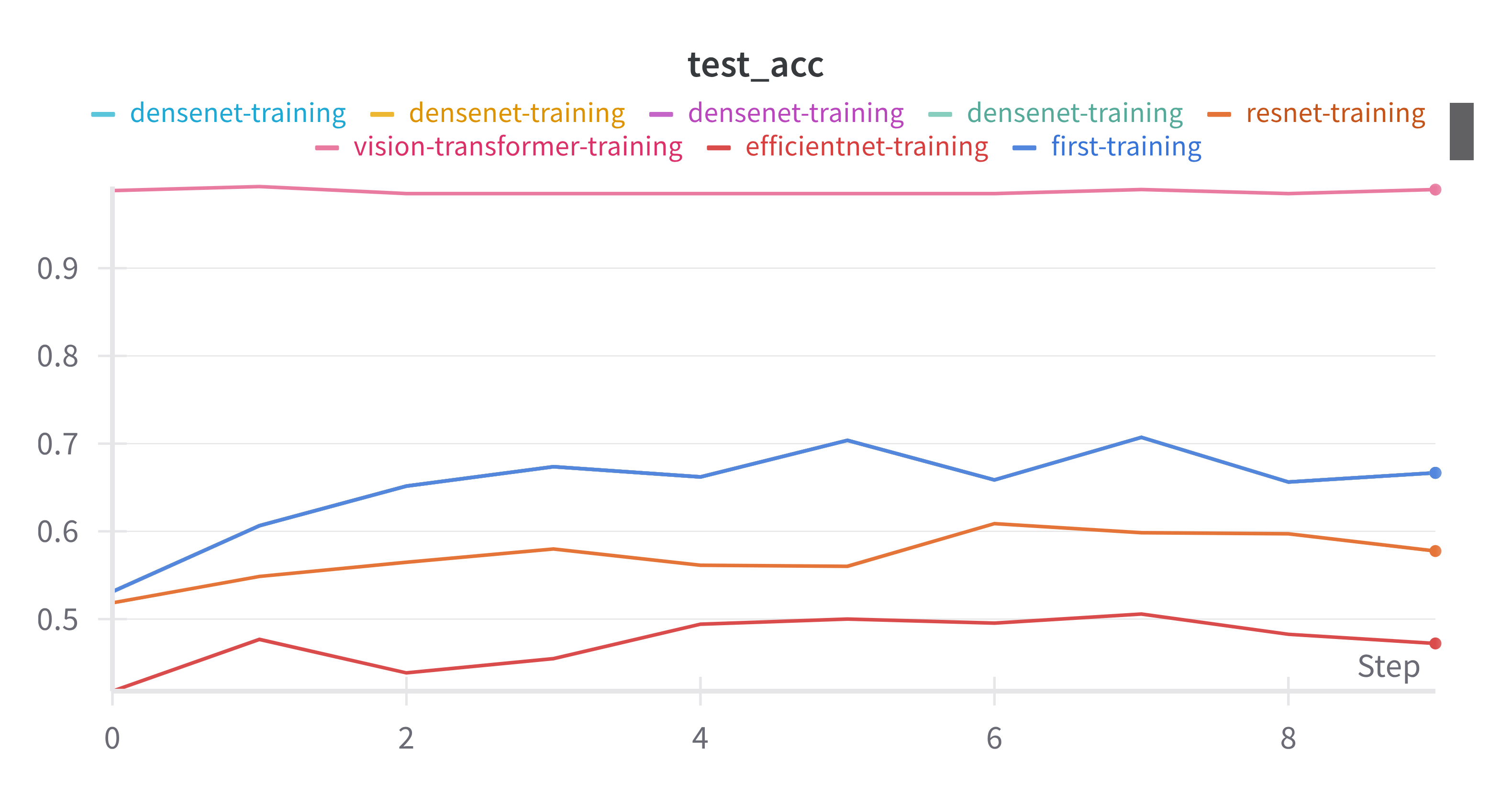}
    \caption{Test accuracy over steps.}
    \label{fig:test-acc}
\end{figure}

\begin{figure}[htbp]
    \centering
    \includegraphics[width=0.5\textwidth]{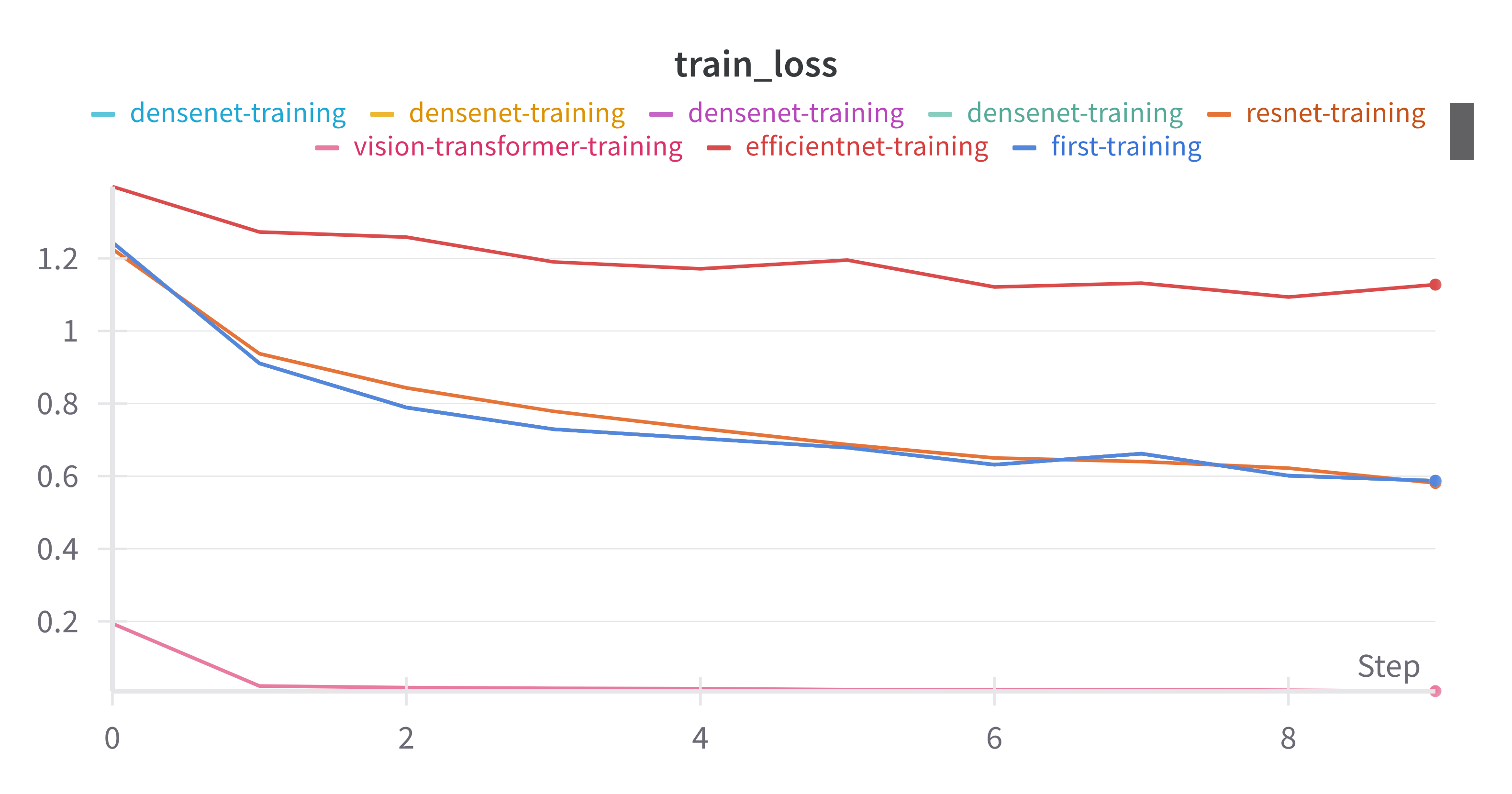}
    \caption{Training loss progression.}
    \label{fig:train-loss}
\end{figure}

\vskip 0.2in

\end{document}